\documentclass[11pt,a4paper]{article}
\usepackage[hyperref]{emnlp2018}
\usepackage{times}
\usepackage{latexsym}

\usepackage{url}
\usepackage{graphicx}
\usepackage{mathtools}
\usepackage{amsmath}
\DeclareMathOperator*{\argmax}{arg\,max}

\aclfinalcopy % Uncomment this line for the final submission

\title{Sentylic at IEST 2018: Gated Recurrent Neural Network and Capsule Network Based Approach for Implicit Emotion Detection }

\author{Prabod Rathnayaka, 
   Supun Abeysinghe, 
  Chamod Samarajeewa \\ 
  \bf{Isura Manchanayake, 
  Malaka Walpola }\\
  Department of Computer Science and Engineering\\
  University of Moratuwa, Sri Lanka\\
  {\tt{\{prabod.14,supun.14,chamod.14,isura.14,malaka\}@cse.mrt.ac.lk}}}

\date{}

\begin{document}
\maketitle

\begin{abstract}
In this paper, we present the system we have used for the Implicit WASSA 2018 Implicit Emotion Shared Task. The task is to predict the emotion of a tweet of which the explicit mentions of emotion terms have been removed. The idea is to come up with a model which has the ability to implicitly identify the emotion expressed given the context words. We have used a Gated Recurrent Neural Network (GRU) and a Capsule Network based model for the task. Pre-trained word embeddings have been utilized to incorporate contextual knowledge about words into the model. GRU layer learns latent representations using the input word embeddings. Subsequent  Capsule Network layer learns high-level features from that hidden representation. The proposed model managed to achieve a macro-F1 score of 0.692.
\end{abstract}

\section{Introduction}

Emotion is a complex aspect of the human behavior which makes the humanity distinguishable from other biological behaviors of creatures. Emotions are typically originated as a response to a situation. Since the emergence of social media, people often express opinions as responses to daily encounters by posting on these platforms. These microblogs contain emotions related to the topics the author have discussed. Thus emotion detection is useful to understand more specific sentiments held by the author towards the discussed topics. Hence this is a challenge with a significant business value.

Emotion analysis can be considered as an extension of sentiment analysis. Even though there has been a notable amount of research in sentiment analysis in the literature, research on emotion analysis has not gained much attention. The related work suggests this task can be handled using emojis or hashtags present in the text i.e. distance supervision techniques \cite{felbo2017using}. However, these features can be unreliable due to noise, thus affect the accuracy of the results.

Although explicit words related to emotions (happy, sad, etc.) in a document directly affect the emotion detection task, other linguistic features play a major role as well. Implicit Emotion Recognition Shared Task introduced in \newcite{Klinger2018x} aims at developing models which can classify a text into one of the emotions; \emph{Anger, Fear, Sadness, Joy, Surprise, Disgust} without having access to an explicit mention of an emotion word. Participants were given a tweet from which any of the above emotion terms or one of their synonyms is removed. The task is to predict the emotion that the excluded word expresses.

E.g.:

\emph{ It's { [\#TARGETWORD\#]} when you feel like you are invisible to others.}

The { [\#TARGETWORD\#]} in the given example corresponds to sadness ("sad").

In this paper, we propose an approach based on Gated Recurrent Units (GRU) \cite{cho2014learning} followed by Capsule Networks \cite{sabour2017dynamic} to tackle the challenge. This model managed to achieve a macro-F1 score of \textbf{0.692} and ranked \textbf{5th} in WASSA 2018 implicit emotion detection task.

\section{Methodology}
We have used a sentence classification model which is based on bidirectional GRUs and Capsule networks. First, the raw tweets are preprocessed, then mapped into a continuous vector space using an embedding layer. Afterward, we used a Bidirectional Gated Recurrent Unit (Bi-GRU) \cite{cho2014learning} layer to encode sentences into a fixed length representation. The fixed length representation is then fed into a Capsule Network \cite{sabour2017dynamic} where it will learn the features and emotional context of the sentences. Finally, the Capsule network is followed by a fully connected dense layer with softmax activation for the classification.

\subsection{Preprocessing}
Microblogs typically contain informal language usages such as short terms, emojis, misspellings, and hashtags. Hence, preprocessing steps should be employed in order to clean these informal and noisy text data. Moreover, efficient preprocessing plays a vital role in achieving a good performance. Ekphrasis tool  \cite{baziotis-pelekis-doulkeridis:2017:SemEval2} is used for initial preprocessing of the tweets. Tweet tokenizing, word normalization, spell correcting and word segmentation for hashtags are done as preprocessing steps.

\subsubsection{Tweet Tokenizing}
Tokenizing is the first and the most important step of preprocessing. Ability to correctly tokenize a tweet directly impacts the quality of a system. Since there is a large variety of vocabulary and expressions present in short texts such as Twitter, it is a challenging task to correctly tokenize a given tweet. Twitter markup, emoticons, emojis, dates, times, currencies, acronyms, censored words (e.g. s**t), words with emphasis (e.g. *very*) are recognized during tokenizing and treated as a separate token.

\subsubsection{Word Normalization}
Upon tokenizing, set of transformations including converting to lowercase and transforming URLs, usernames, emails, phone numbers, dates, times, hashtags to a predefined set of tags (e.g @user1 $\rightarrow$ \textless user\textgreater) are applied. This method helps to reduce the vocabulary size and generalize the tweet.

\subsubsection{Spell Correcting and Word Segmentation}
As the last step in  preprocessing, we apply spell correcting and word segmentation to hashtags. (e.g. \#makeitrain $\rightarrow$ make it rain)

\subsection{Model}
\begin{figure}[h]
    \centering
    \includegraphics[width=0.5\textwidth]{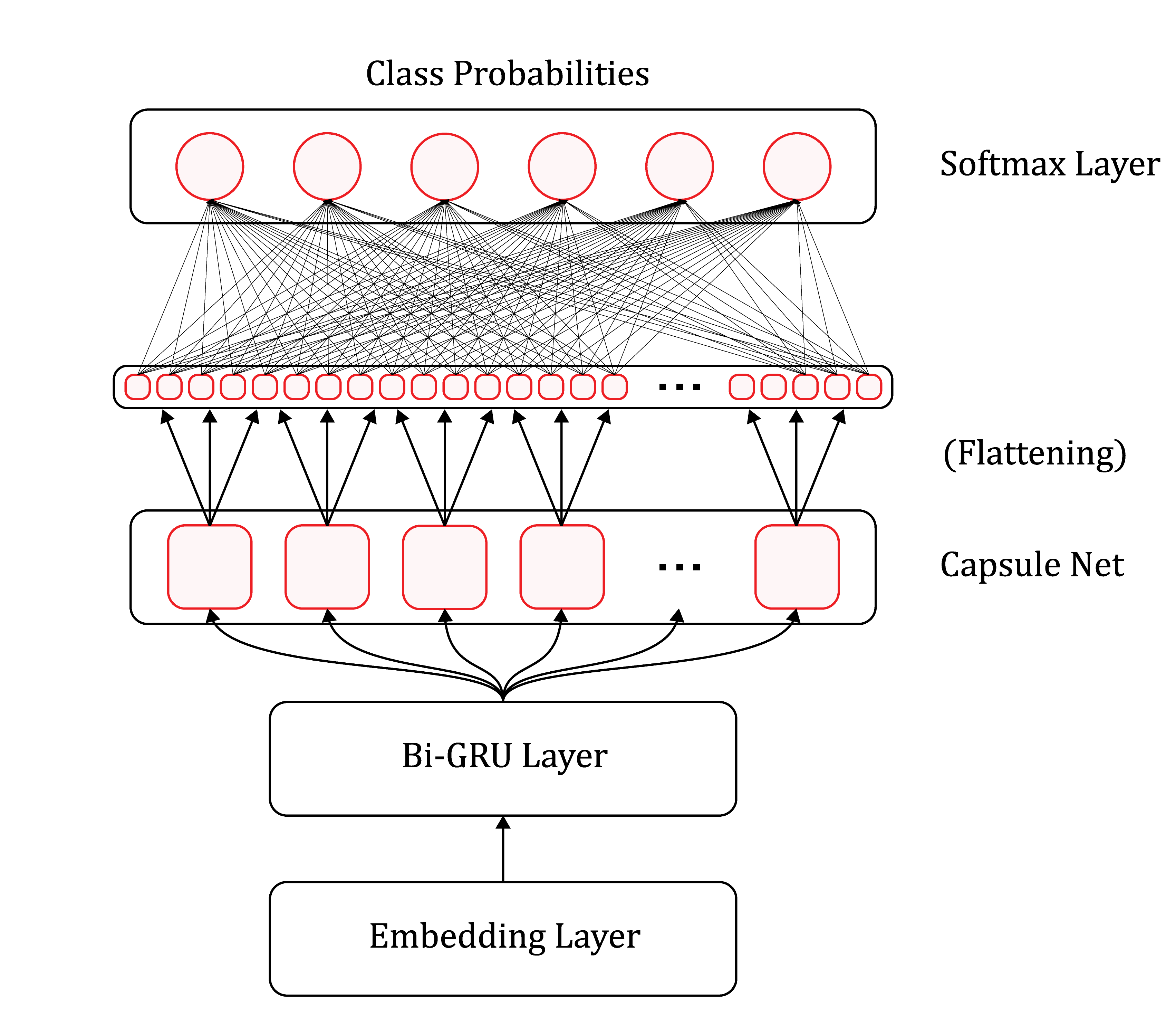}
    \caption{Overall model architecture}
    \label{fig:model}
\end{figure}

An overview of the model is shown in figure \ref{fig:model} and each segment of the model is described in the following sub sections.

\subsubsection{Word Embedding Layer}
Word embedding layer is the first layer of the model. Each token will be mapped into a continuous vector space using a set of pretrained word embeddings. We used 
300 dimensional, pretrained, Word2Vec embeddings introduced in \newcite{mikolov2013efficient}. Given an input tweet, $S = [s_1,s_2,.,s_i,..,s_n]$ where $s_i$ is the token at position $i$, the embedding matrix $W_e$, the output of the embedding layer $X$ is,
\begin{align}
X = SW_e
\end{align}
\subsection{Bidirectional GRU Layer}
The word embedding layer is followed by a bidirectional GRU \cite{cho2014learning} layer. There is a forward GRU ($\overrightarrow{h_t}$) and a backward GRU (($\overleftarrow{h_t}$)) and the latent representation output by the two GRUs is concatenated to get the final output $(\overrightarrow{h_t} ,  \overleftarrow{h_t})$ of the layer. Following set of equations follows the standard notation used in \newcite{cho2014learning}.

\begin{gather}
\begin{align}
r_t &= \sigma(W_{ir} x_t + b_{ir} + W_{hr} h_{(t-1)} + b_{hr}) \\
z_t &= \sigma(W_{iz} x_t + b_{iz} + W_{hz} h_{(t-1)} + b_{hz}) \\
n_t &= \tanh(W_{in} x_t + b_{in} + r_t (W_{hn} h_{(t-1)}+ b_{hn})) \\
h_t &= (1 - z_t) n_t + z_t h_{(t-1)}
\end{align}
\end{gather}

\subsection{Capsule Layer}
Features encoded by the bidirectional GRU layer is then passed to a Capsule Network \cite{sabour2017dynamic}. Capsule Network consists of a set of capsules where each capsule corresponds to a high level feature. Each capsule outputs a vector, of which the magnitude represents the probability of the corresponding feature existence. Following set of equations follows the standard notation used in \newcite{sabour2017dynamic}.

Prediction vector $\hat{u}_{j|i}$ is calculated by multiplying the output $h_i$ from the GRU layer with a weight matrix.
\begin{gather}
\begin{align}
\hat{u}_{j|i} &= W_{ij}h_i
\end{align}
\end{gather}
Total input to a capsule $s_j$ is a weighted sum over all the prediction vectors $\hat{u}_{j|i}$.

\begin{gather}
\begin{align}
s_j &= \sum_{i} c_{ij}\hat{u}_{j|i}
\end{align}
\end{gather}
$c_{ij}$ represents the coupling coefficients found through the iterative dynamic routing.

A non-linear \emph{"Squash"} function is used to scale the vectors such that the magnitude is mapped to a value between 0 and 1.

\begin{gather}
\begin{align}
v_j &= \frac{\Vert{s_j}\Vert^2} {1+\Vert{s_j}\Vert^2} \frac{s_j}{{\Vert{s_j}\Vert}}
\end{align}
\end{gather}
Dynamic Routing process introduced by \newcite{sabour2017dynamic} is used as the routing mechanism between capsules.

\subsection{Classification Layer}
The flattened output from the capsule layer (Say $C$) is fed to a dense layer which has a softmax activation. It outputs a vector of size 6 (number of classes). The values in the vector components are probabilities for the presence of each of the six emotions. The emotion with the highest probability is selected as the output.
\begin{gather}
\begin{align}
Y &= W_{dense}C
\end{align}
\end{gather}
For all $y_i \in Y$, $f_i$ is calculated as follows.  
\begin{gather}
\begin{align}
f_i &= \frac{{e}^{-y_i}}{\sum_{y_j \in Y} e^{-y_j}}
\end{align}
\end{gather}

Then the class with highest $f_i$ is taken as the output.

\begin{gather}
\begin{align}
\text{output} = \argmax_{i} f_i
\end{align}
\end{gather}

\subsection{Regularization}
Gaussian noise is added to both the embedding layer and the softmax classification layer for the purpose of making the model more robust to overfitting. Further, dropout is applied to the Capsule network output and a spatial dropout is applied to the embedding Layer to reduce overfitting.

\section{Experiments and Results}
\subsection{Experimental setup}
\subsubsection{Training}
We used Adam optimizer \cite{kingma2014adam} for optimizing our network with a batch size of 512. Gradient Clipping \cite{pascanu2013difficulty} was employed to address the exploding gradient problem where all the gradients were clipped at 1. Keras \cite{chollet2015keras} was used to develop the model and experiments were done using both Tensorflow \cite{tensorflow2015-whitepaper} and Theano \cite{2016arXiv160502688short} back-ends. Google Colaboratory\footnote{\url{https://colab.research.google.com/}} was used as the runtime environment for training the model.

\subsubsection{Hyper-Parameters}
We have employed Word2Vec \cite{mikolov2013efficient} embeddings of 300 dimensions for the embedding layer. The GRU layer consists of 128 cells for both directions. We have used 16 capsules each with an output size of 32 and 5 routing iterations. Spatial dropout of 0.3 is applied to the embeddings and dropout of 0.25 is applied to the Capsule network. Gaussian noise of 0.1 is added to both the embedding layer and the Capsule network.

\subsection{Results}
We ranked 5th among 30 contestants in the competition. We achieved a macro-F1 score of 0.692 which is 0.155 improvement compared to the baseline model (Maximum Entropy Model using bag of words (BoW) and bigrams). The top-ranked model has a 0.031 improvement compared to our model. Table \ref{results-table} shows the macro-F1 scores of the top 10 competitors and the baseline model.

\begin{table}[t!]
\begin{center}
\begin{tabular}{|l|r|}
\hline \bf Team & \bf Macro-F1 \\ \hline
\bf Amobee & \bf 0.714 \\ 	
IIIDYT & 0.710 \\	
NTUA-SLP & 0.703 \\	
UBC-NLP & 0.693 \\	
\bf Sentylic & \bf 0.692 \\	
HUMIR &	0.686 \\ 	
nlp & 0.685 \\	
DataSEARCH & 0.680 \\ 	
YNU1510 & 0.676 \\ 	
EmotiKLUE &	0.671 \\
\bf Baseline & \bf 0.599 \\
\hline
\end{tabular}
\end{center}
\caption{\label{results-table} Competition results of top 10 competitors and the maximum entropy baseline classifier. }
\end{table}

\section{Analysis}

\subsection{Investigated Approaches}
Recurrent Neural Networks (RNN) \cite{socher2013recursive} based model achieves state-of-the-art in sentence classification tasks. RNNs have the capability to capture sequential features present in sentences. Further, when they are incorporated with attention mechanisms the accuracy of the models increases notably \cite{yang2016hierarchical, tang2015document}. Hence, we have first implemented a model which uses a bidirectional GRU \cite{cho2014learning} layer to learn latent representations followed by a hierarchical attention mechanism. Attention mechanisms have the ability to capture important keywords in sentences and give a higher weight to those words. This is one of the prominent approaches that typically results in a good performance in regular text classification tasks. Table \ref{different-models-table} shows that this approach yielded a reasonable accuracy, yet it was not the best performing approach.

Another approach is to use a Convolution Neural Network (CNN) \cite{kim2014convolutional} layer on top of RNNs instead of attention mechanisms. Intuition is that the CNN layers will act as a different attention mechanism and captures high-level features from the features learned by the below layers. Hence, the second approach we investigated was using CNNs instead of the attention mechanism. As the table \ref{different-models-table} shows, this approach resulted in a slight drop in performance compared to the previous approach.

Our next approach was to use Capsule networks \cite{sabour2017dynamic} instead of Convolution Neural Networks (CNN). Capsule networks have shown promising results in the field of computer vision. \newcite{sabour2017dynamic} argues that it is essential to preserve the hierarchical translational and rotational features of the identified high-level features in order to perform image classification and object detection in the field of computer vision. However, traditional CNNs with max-pooling layers tend to lose this spatial information related to identified features. \newcite{sabour2017dynamic} introduces capsule networks to tackle these issues identified in traditional CNNs. Nonetheless, the usability of Capsule networks has not researched much in the Natural Language Processing (NLP) community. Along the same lines, we can intuitively argue that CNN based models with pooling layers will cause loss of information in text related classification tasks as well. Hence, we have investigated the usability of capsule networks for improving the performance of text classification models. The use of Capsule networks instead of CNNs has improved the performance of the model slightly and assisted in gaining the best performing model.

\begin{table}[t!]
\begin{center}
\begin{tabular}{|l|r|r|r|}
\hline \bf Model & \bf Macro-F1 \\ \hline
GRU + Hierarchical Attention & 0.671 \\ 	
GRU + CNN & 0.657 \\	
\bf GRU + Capsnet & \bf 0.692 \\
\hline
\end{tabular}
\end{center}
\caption{\label{different-models-table} Performance analysis of the best models in each investigated approaches. }
\end{table}

\subsection{Model Architecture Variants}
We have tried several variants of the proposed model. Table \ref{system-variants} shows the performance of each of those variants. We have tried approaches using Long Short Term Memory networks (LSTM) \cite{hochreiter1997long} which is one of the other prominent types of RNNs. However, the results showed a minor drop. Another variant is to use two layers of GRUs instead of using a single layer. Even this approach made the performance of the model slightly lesser. A potential reason for this could be model over-fitting. Using a single GRU layer followed by the Capsnet gave the best performance.

\begin{table}[t!]
\begin{center}
\begin{tabular}{|l|r|r|r|}
\hline \bf Model & \bf Macro-F1 \\ \hline
\bf GRU (1 layer) + Capsnet & \bf 0.692 \\ 	
LSTM (1 layer) + Capsnet & 0.687 \\	
GRU (2 layers) + Capsnet & 0.678 \\
\hline
\end{tabular}
\end{center}
\caption{\label{system-variants} Performance analysis of different variants of the proposed system}
\end{table}

\subsection{Analysis on Predictions}

\begin{table}[t!]
\begin{center}
\begin{tabular}{|l|l|l|l|}
\hline \bf Emotion & \bf Precision & \bf Recall & \bf F1-Score \\ \hline	
Anger & 0.631 & 0.614 & 0.622\\
Disgust & 0.689 & 0.687 & 0.688\\
Fear & 0.728 & 0.731 & 0.730\\
Joy & 0.803 & 0.774 & 0.788\\
Sad & 0.682 & 0.655 & 0.668\\
Surprise & 0.625 & 0.689 & 0.656\\
\hline
\bf Micro Avg. & 0.694 & 0.694 & 0.694\\
\bf Macro Avg. & 0.693 & 0.692 & 0.692\\
\hline
\end{tabular}
\end{center}
\caption{\label{f1-results-table} Precision, recall and F1-score of each class in test set using our proposed model. }
\end{table}

Table \ref{f1-results-table} shows the performance of the proposed model for each class. As evident from the results, anger shows a significantly lower F1-score. Other emotions show similar results whereas joy stands out with a notably higher F1-score. Anger has been misclassified as sad in several examples.

e.g.- \textit{Girls will get [{\#TARGETWORD\#}] that her man cheated with an ugly girl more than the fact he actually cheated.}

In the above example, it is unclear whether the emotion is anger or sadness. Such ambiguity of anger has affected the reduction of F1-score values. There were few other similar cases where it is challenging even for humans to clearly discriminate emotions due to nuance nature of emotions expressed. 

\section{Conclusion}
WASSA 2018 Implicit Emotion Shared Task \cite{Klinger2018x} introduces a task to predict the emotion of a tweet of which the explicit mentions of emotion terms have been removed. We have experimented with several deep learning based approaches to tackle this task. We have used pre-trained Word2Vec embeddings. All the approaches we tried utilize an initial GRU layer which learns latent representations from the input word embeddings. Different alternative methods have been investigated for the subsequent layer. These methods include attention layer, CNN layer, and Capsnet layer. Model with the Capsnet layer achieved the best results among the experimented alternatives.
Potential future work includes investigating the possibility of using Capsule networks for other tasks in Natural Language Processing, especially where CNNs are involved. Another line of future work could be to follow the approach mentioned in \newcite{felbo2017using} and apply transfer learning on the model trained using this semi-automatically annotated dataset to test on human annotated datasets such as \newcite{mohammad2018semeval}.

\bibliography{emnlp2018}
\bibliographystyle{acl_natbib_nourl}

\end{document}